\documentclass[12pt,oneside]{article} 

\usepackage{booktabs}

\usepackage[comma]{natbib}
\bibpunct[, ]{(}{)}{;}{a}{}{,}

\usepackage{authblk}

\usepackage[colorlinks, linkcolor = black, citecolor = black, filecolor = black, urlcolor = black]{hyperref}
\usepackage{graphicx}
\usepackage{subfig}
\usepackage{amsmath,amsthm}
\usepackage{amssymb}
\usepackage{titlesec}
\usepackage{tikz}
\usepackage[text={6.3in,9.5in},a4paper]{geometry}
\usepackage{setspace}
 \usepackage{color}
 \usepackage{colortbl}
 \usepackage{linguex}
 \usepackage{arydshln}
 \usepackage{xcolor}

\usepackage{fancyhdr}
\usepackage{wasysym}

\usepackage[labelfont={normalsize,bf},font={normalsize,singlespacing},labelsep=period,justification=centering]{caption}

\fancypagestyle{plain}{%
\fancyhead{} 
}
\usepackage{color}

\definecolor{blue}{gray}{0}

\begin{document}

\titleformat{\section}
  {\normalfont\rmfamily\bfseries\color{black}}
  {\thesection}{1em}{}

\titleformat{\subsection}
  {\normalfont\rmfamily\bfseries\color{black}}
  {\thesubsection}{1em}{}


\begin{titlepage}
\title{\Large A Means-End Account of Explainable Artificial Intelligence} 

\author{\em Oliver Buchholz\thanks{University of T\"ubingen, Cluster of Excellence ``Machine Learning: New Perspectives for Science'', T\"ubingen, Germany. E-mail: \href{mailto:oliver.buchholz@uni-tuebingen.de}{\texttt{oliver.buchholz@uni-tuebingen.de}}\\I would like to thank audiences in T\"ubingen and Frankfurt a. M., in particular Kate Vredenburgh and David Danks, for helpful comments on a talk based on this article. I would also like to thank Sara Blanco, Karoline Reinhardt as well as Thomas Grote for helpful feedback and Wolfgang Spohn for very valuable comments on an earlier draft. Finally, I would like to thank Eric Raidl for his guidance and continuing support during the entire process that led up to this article.\\ My research was funded by the Baden-W\"urttemberg Foundation (program ``Verantwortliche K\"unstliche Intelligenz'') as part of the project AITE (Artificial Intelligence, Trustworthiness and Explainability).} \mbox{     }}

\maketitle

\thispagestyle{empty}
\begin{abstract}
{\footnotesize
\noindent Explainable artificial intelligence (XAI) seeks to produce explanations for those machine learning methods which are deemed opaque. However, there is considerable disagreement about what this means and how to achieve it. Authors disagree on \emph{what} should be explained (topic), \textcolor{blue}{\emph{to whom} something should be explained (stakeholder)}, \emph{how} something should be explained (instrument), and \emph{why} something should be explained (goal). In this paper, I employ insights from means-end epistemology to structure the field. According to \emph{means-end epistemology}, different means ought to be rationally adopted to achieve different epistemic ends. Applied to XAI, different topics, \textcolor{blue}{stakeholders,} and goals thus require different instruments. I call this the \emph{means-end account} of XAI. The means-end account has a descriptive and a normative component: on the one hand, I show how the specific means-end relations give rise to a taxonomy of existing contributions to the field of XAI; on the other hand, I argue that the suitability of XAI methods can be assessed by analyzing whether they are prescribed by a given topic, \textcolor{blue}{stakeholder,} and goal.
}
\end{abstract}


\end{titlepage}


\newpage
\pagestyle{fancy}
\setcounter{page}{2}

\section{Introduction}

Methods of machine learning (ML) are gaining relevance in a variety of domains. They are employed to navigate self-driving vehicles \citep{Bojarski.2016}, support medical diagnosis \citep{Esteva.2017}, and to detect objects ranging from particles at the subatomic level \citep{Baldi.2014} to exoplanets \citep{ForemanMackey.2015}.\footnote{I use the term `ML method' to refer to some general methodology, e.g., deep learning, and distinguish it from the term `ML model' by which I refer to a specific model, e.g., a deep neural network for which the weights have already been determined. Thus, the term `ML method' explicitly includes the learning process, while the term `ML model' only refers to the learned function.}

At the same time, it is widely acknowledged that their very nature makes ML methods essentially `black boxes' to human agents. Both their specific functioning and their complexity, especially in the field of deep learning, are inherently opaque or beyond human grasp (\citeauthor{Burrell.2016} \citeyear{Burrell.2016}, \citeauthor{Creel.2020} \citeyear{Creel.2020}). Thus, frequently, laymen and experts alike do not know how these methods work, why they are successful, or why they fail. This observation led to the widely shared belief that ML methods should be made explainable and sparked research in the field of explainable artificial intelligence (XAI). Since the `X' in `XAI' stands for `explainable', it seems straightforward and uncontroversial to summarize the central aim of the field plain and simple as follows: XAI seeks to provide instruments that produce explanations of ML methods.\footnote{To be precise, this applies to the field insofar it is concerned with the application of XAI techniques to opaque ML methods, which is the focus of this paper. It does not directly apply to that part of XAI that seeks to develop and apply inherently interpretable methods, thereby avoiding the use of opaque ML methods from the outset \citep{Rudin.2019}.}

Yet on the one hand, there is profound conceptual disagreement regarding the precise \emph{meaning} of this aim (\citeauthor{Lipton.2018} \citeyear{Lipton.2018}): what is it that explanations of ML methods should achieve? Suggestions in the literature comprise the straightforward requirements of \emph{explainability} and \emph{interpretability} (\citeauthor{DoshiVelez.2017} \citeyear{DoshiVelez.2017}, \citeauthor{Erasmus.2020} \citeyear{Erasmus.2020}), the more intuitive \emph{understandability} \citep{Paez.2019} as well as the larger category of \emph{transparency} (\citeauthor{Gunther.2021} \citeyear{Gunther.2021}, \citeauthor{Zerilli.2019} \citeyear{Zerilli.2019}). Others argue that all these notions lack precise definitions altogether and only hide the real goals such as safety of or non-discrimination by ML methods \citep{Krishnan.2020}.

On the other hand, the literature also diverges with respect to the \emph{instruments} that are deemed appropriate to achieve the aim of XAI. Popular approaches range from local approximations of complex models \citep{Ribeiro.2016} to visual \citep{Xu.2015}, textual \citep{Hendricks.2016}, counterfactual-based \citep{Wachter.2018} as well as attribution-based explanations \citep{Lundberg.2017}. Overall, it is unclear whether the field of XAI is scattered into various disconnected subprojects or whether there is a common structure that is shared by the variety of approaches pursued in the literature.

In this article, I propose an account that structures the field. To do so, I employ insights from means-end epistemology.\footnote{The idea of exploiting existing research from philosophy and the social sciences when analyzing XAI is put forward, e.g., in \citet{Miller.2019}.} \emph{Means-end epistemology} takes epistemology to be a normative discipline; it is based on the principle of instrumental rationality (\citeauthor{Huber.2021} \citeyear{Huber.2021}: 1). Rational agents are assumed to have certain epistemic ends (e.g., the end to hold true beliefs), and such agents ought to adopt certain means (e.g., to base their beliefs on total evidence) if and only if they further these epistemic ends (\citeauthor{Huber.2021} \citeyear{Huber.2021}: 125, \citeauthor{Schulte.1999} \citeyear{Schulte.1999}: 7). Providing an explanation is such an epistemic end, because it relates to the epistemic state of the agent receiving the explanation.\footnote{\textcolor{blue}{This is also a well-established result in psychology as reported, for instance, by \citet{Keil.2006} or \citet{Lombrozo.2006}.}} Thus, when trying to provide instruments that produce explanations of ML methods, the field of XAI essentially seeks to provide the means to further an epistemic end. Consequently, means-end epistemology is a very suitable framework to analyze XAI. I call this the \emph{means-end account of XAI}. The account's motivation is based on the fundamental observation that means-end relations are crucial to the field of XAI. Indeed, as I will point out, researchers explicitly or implicitly specify them in their contributions. I will show how these means-end relations can be specified on a fine-grained level by distinguishing \emph{what} should be explained (topic), \textcolor{blue}{\emph{to whom} something should be explained (stakeholder)}, \emph{why} something should be explained (goal), and \emph{how} something should be explained (instrument).

I will argue that overall, the means-end account has several important consequences. First, it explains why disagreement arises in the field: the divergence in methods of XAI follows from the disagreement on fine-grained ends being pursued. Second, it unifies and structures the field: there is a shared methodology of developing appropriate means for given ends. Third, this structure has a descriptive component: different authors specify different topics\textcolor{blue}{, address different stakeholders, and have different goals}. Thus, they pursue different ends and different means are appropriate to achieve them. This gives rise to a taxonomy that classifies existing contributions to the field along the specific means-end relations that are considered. Fourth, the means-end structure also has a normative component: according to means-end epistemology, different means ought to be rationally adopted to achieve different epistemic ends. Therefore, the fine-grained ends of an explanation normatively constrain the set of admissible means to achieve it. The means-end account thus reveals how the suitability of particular instruments of XAI is prescribed by the ends for which an explanation is sought. I argue that this analysis gives rise to a normative framework that can be used to assess the suitability of XAI techniques.

\textcolor{blue}{Consequently, this article extends that strand of the literature that emphasizes the relevance of pragmatic considerations in the context of XAI.\footnote{\textcolor{blue}{Examples include \citet{Besold.2018}, \citet{Nyrup.2022}, and \citet{Paez.2019}.}} Indeed, I show that pragmatic considerations are relevant, especially in what concerns the topics, stakeholders, and goals. However, by focusing on the principle of instrumental rationality as a cornerstone of epistemic normativity, I additionally provide a normative justification for why pragmatic considerations \emph{ought} to be relevant. Furthermore, while I do not claim to be the first developing a normative framework for XAI, this article is the first to thoroughly base such a framework on an underlying epistemological theory.\footnote{\textcolor{blue}{Other normative frameworks for XAI have been proposed, for instance, by \citet{Langer.2021}, \citet{Mohseni.2021}, and \citet{Zednik.2019}.}}}

The remainder of this article is structured as follows: In Section~\ref{sec:prelims}, I outline the main characteristics of XAI by providing concrete examples from the field. I also give an introduction to means-end epistemology. In Section~\ref{sec:meansend}, I argue that problems of XAI can be framed as problems of means-end epistemology. I thereby establish the means-end account of XAI and discuss both its normative and its descriptive component. \textcolor{blue}{In Section~\ref{sec:extension}, I discuss an extension of the account} before concluding in Section~\ref{sec:conclusion}.

\section{Preliminaries}
\label{sec:prelims}

This section outlines the main characteristics of XAI by examining two examples from the field. It also gives a short introduction to those aspects of means-end epistemology that are relevant for the rest of the paper.

\subsection{Explainable Artificial Intelligence}
\label{subsec:xai}

XAI is commonly employed to eliminate the inherent opacity of many ML methods and models. In a first step, there is usually some opaque ML model, for instance a deep neural network, whose opacity should be (at least partly) eliminated. In a second step, some XAI technique is applied to the opaque ML model to achieve that elimination. Yet as straightforward as it may seem, there is considerable disagreement about what `eliminating opacity' amounts to. This disagreement is perhaps best illustrated by outlining two of the most well-known methods of XAI. 

First, consider local interpretable model-agnostic explanations (LIME) \citep{Ribeiro.2016}. Explanations produced by LIME are \emph{local} in the sense that they explain ML predictions only within a specific part of the data. To do so, LIME approximates the possibly highly complex ML model by a simpler model around the specific prediction(s) of interest. Thus, a complex decision surface might be approximated locally by a linear function. This makes the overall ML model locally \emph{interpretable}, thereby revealing the model's structure within a specific part of the data. Since LIME is \emph{model-agnostic}, it is applicable to a wide range of ML methods.

Second, consider counterfactual explanations (CFE) \citep{Wachter.2018}. 
They are usually provided as verbalized counterfactual statements that take the following form: ``You were denied a loan because your annual income was \pounds 30,000. If your income had been \pounds 45,000, you would have been offered a loan.'' Thus, CFE provide reasons for a given decision and seem to specify how to reach a desired result in the future.\footnote{I use the rather cautious `seem', since this view is controversial. See Section~\ref{subsec:normative} for a brief overview about the debate in XAI. See also \citet{Raidl.2022} as well as references therein for the general philosophical debate about the relation of counterfactuals and `because'-statements.} As becomes obvious, CFE exclusively focus on a model's input-output relation: they explain why a certain input led to a certain output ``without opening the black box'', that is, without considering a model's internal functioning. They justify the model's output by giving reasons rather than by detailing the decision rule that led to the output.

Overall, the deliberately simplistic sketch of two of the most popular methods in XAI illustrates the observation from above: that instruments of XAI differ widely and that there is no consensus on which instruments are appropriate to use. This raises the question whether there is nevertheless something that XAI techniques, ranging from LIME to CFE, have in common. As a first step, let us try to distill an overall aim of XAI, general enough to be shared by LIME, CFE, and other XAI techniques as different as they may be: what is it, that XAI tries to achieve?

First, XAI develops \emph{instruments}, for instance, LIME or CFE. Second, these instruments are meant to produce \emph{explanations} of the ML methods to which they are applied. 
Thus at the highest level, the aim of XAI might be summarized by stating that XAI seeks to

\ex. Provide instruments that produce explanations of ML methods. \label{eq:goal}

Clearly, the term `explanations' in \ref{eq:goal} is dangerously loaded, both within XAI and within philosophy. In this article, I use it to designate the different kinds of outputs that are produced by XAI techniques, regardless of whether they qualify as a proper explanation according to some definition or not.  At this stage, my sole aim is to ensure a maximally general formulation of the high-level goal of XAI that is meant to be as uncontroversial as possible. I will return to it below when putting forward the main argument.


\subsection{Means-End Epistemology}
\label{subsec:mee}

The second building block of the account that I propose in this paper is means-end epistemology. It takes epistemology to be a normative discipline. In this context, normativity has an epistemic rather than a moral meaning. Contrary to moral normativity that is concerned with what one ought to \emph{do}, epistemic normativity is concerned with what one ought to \emph{believe}. In the case of means-end epistemology, the central normative criterion is spelt out by the principle of instrumental rationality (\citeauthor{Huber.2021} \citeyear{Huber.2021}, \citeauthor{Schulte.1999} \citeyear{Schulte.1999}):

\ex. Given an epistemic end, certain means ought to be adopted if and only if they further the epistemic end. \label{eq:mee}

Thus, agents are assumed to have certain epistemic ends and the principle of instrumental rationality requires them to adopt those means that are appropriate to achieve the given end(s).\footnote{Note, that \ref{eq:mee} expresses a \emph{hypothetical} rather than a \emph{categorical} imperative. It is controversial whether `ought' needs to be replaced by `should' in this case (\citeauthor{Finlay.2009} \citeyear{Finlay.2009}, \citeauthor{Foot.1972} \citeyear{Foot.1972}, \citeauthor{Huber.2021} \citeyear{Huber.2021}: Ch. 5.4). Since engaging with this debate is beyond the scope of this text, my use of `ought' is only meant to emphasize the pronounced normative perspective of means-end epistemology.} For instance, consider Anne, who pursues the epistemic end of holding only true beliefs. She falsely believes that she was denied a loan because her annual salary is too low. Now suppose that she receives the (true) information that, in fact, she was denied the loan because her loan application contained transposed digits for her annual salary. According to means-end epistemology, Anne---if rational---ought to adopt means that lead to a revision of her initial and false belief in favor of the true belief that her loan application contained transposed digits for her annual salary, since these means would further her epistemic end of holding only true beliefs.

Equating epistemic normativity with instrumental rationality and expressing it in terms of means-end relationships as in \ref{eq:mee} involves several aspects that are important for the subsequent discussion. First, different ends call for different means. Thus, while the adoption of certain means might be rational for agents with certain epistemic ends, agents with different ends need not or even \emph{ought} not adopt the same means (\citeauthor{Schulte.1999} \citeyear{Schulte.1999}: 26). This is a consequence of the normative perspective adopted by means-end epistemology.

Second, note that `certain means' in \ref{eq:mee} generally refers to a set of means, $M$, rather than to one particular means, $m \in M$. Thus, different ends may call for different \emph{sets} of means, $M$ and $M'$, but the latter can be partially overlapping, such that $M \cap M' \neq \emptyset$. So on the one hand, \emph{different} ends might be achieved by the \emph{same} particular means $m$. On the other hand, the \emph{same} end might be achieved by \emph{different} particular means, $m$ and $m'$.

Third, epistemic ends come in different levels of granularity. For instance, \citet{Schulte.1999} conducts a means-end analysis of inductive inference. His starting point is the epistemic end of beliefs converging to the truth. Subsequently, however, further details are added to this initial end by requiring the convergence to possess certain properties, most importantly to evolve fast and with minimal retractions. Clearly, the latter ends are more fine-grained than the former, thereby allowing to identify a more specific set of means that ought to be rationally adopted.

Fourth, note that the means-end statement in \ref{eq:mee} has the logical form of a biconditional and is therefore equivalent to a conjunction of two conditional statements pointing in two opposite `directions'\footnote{This is due to the elementary fact that for two propositions $P$ and $Q$, the biconditional $P \leftrightarrow Q$ holds if and only if the conjunction of the single conditionals $P \rightarrow Q$ and $Q \rightarrow P$ holds.}: according to one direction, \emph{if} certain means ought to be adopted, \emph{then} they further the epistemic end; according to the other and, arguably, more natural direction, \emph{if} they further the epistemic end, \emph{then} certain means ought to be adopted. The importance of taking both directions into account and thus preserving the logical form of the biconditional will become important below.

\section{A Means-End Account of XAI}
\label{sec:meansend}

This section establishes what I shall call the \emph{means-end account} of XAI. It also discusses the account's descriptive and normative component.

\subsection{Establishing the Account}
\label{subsec:argument}

Recall the overall goal of XAI that I proposed in \ref{eq:goal}: XAI seeks to provide instruments that produce explanations of ML methods. On the one hand, this means that XAI clearly seeks to affect the epistemic states of individuals, be it their beliefs about or their understanding of ML methods---the observation does not hinge on a particular epistemological concept such as belief, knowledge, or understanding being affected. Consequently, it is fair to say that by producing explanations, XAI pursues an epistemic end. On the other hand, XAI seeks to provide methods to achieve this epistemic end. Put differently, XAI seeks to find appropriate means to further a given epistemic end. Taken together, these aspects reveal that

\ex. Problems of XAI can be framed as problems of means-end epistemology. \label{eq:xai_mee}

One might object that it is somewhat circular to argue for \ref{eq:xai_mee} based on the high-level goal of XAI proposed in \ref{eq:goal}, since the latter could have been purposefully designed for precisely this argument. However, as mentioned above, \ref{eq:goal} is simply formulated on the highest level possible so as to give rise to a maximally broad and uncontroversial statement.

Let me illustrate the insight in \ref{eq:xai_mee} using the examples from Section~\ref{subsec:xai}. First, consider LIME: as outlined above, LIME seeks to explain an individual prediction or several predictions in a specific region of the data. This constitutes the epistemic end that is pursued in this case. The means proposed to achieve this end are local approximations to a possibly complex ML model. Second, consider CFE: here, the epistemic end is to explain the input-output relation of a model. The means proposed in this case are counterfactual statements like the one about loan denial given above. So indeed, means-end relations seem useful and easily applicable to analyze XAI techniques.

However, when framing problems of XAI as problems of means-end epistemology, it might seem trivial to proceed by simply specifying means and ends corresponding to existing methods like LIME or CFE. The more important question is how means-end relations are determined in practice, for instance, in the development of new methods. Analyzing the means-end relations governing XAI more closely reveals that researchers either explicitly or implicitly answer \textcolor{blue}{a variety of} different questions: \emph{what} should be explained, \textcolor{blue}{\emph{to whom} it should be explained,} and \emph{how} it should be explained. I argue that the answers that are specified for each of the questions determine the relevant means-end relations.

Answering the question as to \emph{what} should be explained determines the \emph{topic} of the explanation. At first blush, this might seem unnecessary: when trying to produce an explanation of some ML method, the topic should obviously be the ML method itself. Yet the topic can also be spelled out at a more fine-grained level as a particular aspect of an ML method and, indeed, it regularly is. This is shown in Table~\ref{tab:test}: in the case of LIME, the focus is on explanations for predictions within a specific part of the data while in the case of CFE, it is on the input-output relation of an ML model.

\textcolor{blue}{Answering the question \emph{to whom} something should be explained determines the relevant \emph{stakeholder} who is asking for and receiving the explanation. This is important, since there is a variety of different stakeholders in the `ML ecosystem'. The latter term was coined by \citeauthor{Tomsett.2018} (\citeyear{Tomsett.2018}) who distinguish creators (agents that create an ML system), operators (agents interacting directly with an ML system), and executors (agents making decisions informed by an ML system) as well as decision-subjects (agents affected by ML-based decisions) and data-subjects (agents whose personal data has been used to train the system).\footnote{\textcolor{blue}{This is only one possible classification of stakeholders and I will use it here as a reference point. For another classification, see, for instance, \citeauthor{Preece.2018} (\citeyear{Preece.2018}).}} Clearly, this variety of stakeholders suggests a variety of cognitive abilities, previous knowledge, or interests across the different agents. It has therefore been pointed out repeatedly that different stakeholders have different explanatory requirements (\citeauthor{Langer.2021} \citeyear{Langer.2021}, \citeauthor{Mohseni.2021} \citeyear{Mohseni.2021}, \citeauthor{Zednik.2019} \citeyear{Zednik.2019}).\footnote{Recently, it has been argued that, in addition, what matters is the relation between the stakeholder receiving and the stakeholder providing the explanation \citep{Bordt.2022}.} This is reflected in practice: for instance, LIME is explicitly meant to provide explanations to the users of an ML system while CFE are meant to do the same for data-subjects (\citeauthor{Ribeiro.2016} \citeyear{Ribeiro.2016}: 1135, \citeauthor{Wachter.2018} \citeyear{Wachter.2018}: 843).\footnote{\textcolor{blue}{However, note that there is no standard terminology for the stakeholders in the ML ecosystem yet: the users mentioned by \citeauthor{Ribeiro.2016} (\citeyear{Ribeiro.2016}: 1135) are likely to be either operators or executors as defined in the classification by \citet{Tomsett.2018}. Similarly, based on the remarks by \citeauthor{Wachter.2018} (\citeyear{Wachter.2018}), it seems likely that CFE should rather be provided to \emph{decision}- rather than to \emph{data}-subjects as defined in the classification by \citet{Tomsett.2018}. Thus, to avoid ambiguities, Table~\ref{tab:test} was constructed using the latter classification.}}}

\textcolor{blue}{The previous paragraphs reveal that the specification of a topic and the relevant stakeholders further characterizes what aspects should be taken into account when an XAI technique produces an explanation. This determines an epistemic end that is one component of the epistemic means-end relations governing XAI. Clearly then, the corresponding means are the other component. They are determined by answering the question as to \emph{how} an explanation should be achieved, for this specifies} the \emph{instruments} deemed appropriate to achieve that explanation at hand. As reflected by the high-level aim in \ref{eq:goal}, this is the main occupation of methodological research in XAI: to develop and improve instruments that produce explanations of ML methods. In the case of LIME, local approximations of complex models are the instruments or means proposed to attain the given epistemic end, while CFE seek to attain the given epistemic end via counterfactual statements (see Table~\ref{tab:test}).


\begin{table*}
\footnotesize
\centering
\caption{Possible determination of fine-grained means-end relations for LIME and CFE.}
\begin{tabular}{lp{4cm}p{3.5cm}p{4cm}}
\toprule
& \multicolumn{2}{c}{Epistemic End} & \multicolumn{1}{c}{Means}\\
\cmidrule(lr){2-3} \cmidrule(l){4-4} 
& \emph{What?} & \emph{To whom?} & \emph{How?}\\
Example & (topic) & (stakeholder) & (instrument)\\
\midrule
LIME & prediction(s) in specific part of the data & operators; executors & local approximation to complex model\\
&&&\\
CFE & input-output relation & decision-subjects & counterfactual statements\\
\bottomrule
\end{tabular}
\label{tab:test}
\end{table*}

Although the exact characterization of LIME and CFE presented in Table~\ref{tab:test} might be debatable, the overall framework discussed so far clearly gives rise to a situation that is considerably more fine-grained than the high-level aim of XAI formulated in \ref{eq:goal}: epistemic ends can be split into topic and stakeholder, instruments constitute the corresponding means to achieve given ends. \textcolor{blue}{Thus, the high-level aim can be reformulated by stating that XAI seeks to}

\ex.[(1')] \textcolor{blue}{Provide instruments that produce explanations of topic $t$ for stakeholder $s$.} \label{eq:ext1}

On the one hand, this goes slightly beyond standard means-end epistemology that takes epistemic ends to come in different levels of granularity, yet also takes them as rather monolithic and not separable into further constituents. On the other hand, it hints at a possible strategy to determine the different levels of granularity of epistemic ends in practice: there is a coarse-grained level as expressed in \ref{eq:goal}, stating that an explanation of an ML method should be produced; yet there is a sequence of more fine-grained levels \textcolor{blue}{as expressed in (1'),} at which explanations of specific parts of said ML method should be produced for specific \textcolor{blue}{stakeholders}. Stating topic, \textcolor{blue}{stakeholder}, and instrument with increasing precision allows to move from the coarse- to the fine-grained level. As I will show in the following, this kind of analysis has a descriptive and a normative component.

\subsection{The Account's Descriptive Component}
\label{subsec:descriptive}

The means-end account proposed in the previous section has a rather straightforward descriptive component: as illustrated in Table~\ref{tab:test}, it allows to \emph{describe} the means-end considerations of XAI researchers by stating what topic and \textcolor{blue}{stakeholder} constitute their epistemic end and what instruments they propose as means to attain it. As two further examples will show, this is a useful framework for structuring the field of XAI.

First, consider a technique proposed by \citeauthor{Hendricks.2016} (\citeyear{Hendricks.2016}). 
It is meant to produce explanations of how some visual input to an ML-based image classification system, that is, an image, leads to a specific output, that is, a particular classification. Their solution is to generate textual explanations that describe those parts of the visual input that distinguish it from images in other categories.

Second, consider a technique proposed by \citeauthor{Kim.2018} (\citeyear{Kim.2018}). It is meant to produce explanations for the internal state of an ML system that is controlling an autonomous vehicle. Their solution is to generate textual explanations derived from the visual attention of the ML system that controls the vehicle.

Exploiting the means-end account's descriptive component, it is straightforward to analyze both techniques just like LIME and CFE above. As for the \emph{topic}, \citeauthor{Hendricks.2016} (\citeyear{Hendricks.2016}) focus on the relationship between an input to the image classification system and its corresponding output. In fact, they make their specification of the topic explicit by stating that they concentrate on ``\emph{justification} explanation systems producing sentences detailing how visual evidence [i.e., some input] is compatible with a system output'' (\citeauthor{Hendricks.2016} \citeyear{Hendricks.2016}: 3). Furthermore, they contrast their approach with ``\emph{introspective} explanations'' that focus on the internal functionality of a system. The latter, however, are the explicit topic defined by \citeauthor{Kim.2018} (\citeyear{Kim.2018}: 564) who state that they aim at providing ``explanations that are based on the system's internal state''. \textcolor{blue}{As for the \emph{stakeholder} who should receive the explanation, both techniques are meant to address either the operators or the executors of an ML system.}\footnote{\textcolor{blue}{To be precise, both \citet{Hendricks.2016} and \citet{Kim.2018} mention (end-)users as the relevant stakeholders, which in terms of the classification by \citet{Tomsett.2018} are either operators or executors.}}

Finally, as for the \emph{instruments}, \citeauthor{Hendricks.2016} (\citeyear{Hendricks.2016}) propose a technique that consists in describing those parts of input images that are decisive for their classification. To use the authors' terminology, they propose to generate textual explanations that describe properties that are both ``image-relevant'' in the sense that the property is really contained in the input at hand and ``class-discriminative'' in the sense that the property is relevant for distinguishing between different categories (\citeauthor{Hendricks.2016} \citeyear{Hendricks.2016}: 4). For the particular application that the authors investigate, the classification of birds, this leads to explanations like ``[t]his is a \emph{Western Grebe} because this bird has a long white neck, pointy yellow beak and red eye'' (\citeauthor{Hendricks.2016} \citeyear{Hendricks.2016}: 2). The explanation mentions the output of the image classification system, `Western Grebe', as well as specific properties that occur in the input image and that distinguish the Western Grebe from other, similar looking birds. 

\citeauthor{Kim.2018} (\citeyear{Kim.2018}) propose another instrument that is known as visual attention heatmapping to achieve their epistemic end (\citeauthor{Kim.2017} \citeyear{Kim.2017}, \citeauthor{Xu.2015} \citeyear{Xu.2015}). This means that their technique to achieve explanations relies on analyzing the visual attention of the ML system controlling the autonomous vehicle. More precisely, the system is continuously confronted with an input of traffic scenes consisting of dashcam images and sensor measurements such as the vehicle's speed. The authors employ a so-called attention model to extract salient features of this input that the system `looks at' and that it uses for its decisions regarding the vehicle's acceleration as well as its change of course. Subsequently, textual descriptions for these salient features are generated, leading to explanations for the vehicle's behavior such as: ``The car is driving forward + because there are no other cars in its lane'' (\citeauthor{Kim.2018} \citeyear{Kim.2018}: 564). Although this explanation resembles the ones produced by the technique of \citeauthor{Hendricks.2016} (\citeyear{Hendricks.2016}), it is generated by a strictly different approach: whereas explanations by \citeauthor{Hendricks.2016} (\citeyear{Hendricks.2016}) are generated conditional on visual properties of the ML system's \emph{input} (e.g., `long white neck', `pointy yellow beak', `red eye'), explanations by \citeauthor{Kim.2018} (\citeyear{Kim.2018}) rely on attention heatmaps and are thus generated conditional on \emph{internal states} of the ML system. To emphasize this distinction, I refer to the former approach as `textual explanations' and to the latter as `verbalized attention heatmapping'.

The new examples just discussed can be used to extend Table~\ref{tab:test}, giving rise to the situation in Table~\ref{tab:taxonomy}: four XAI techniques are described based on the specific means-end considerations by which they are governed. This is useful for at least two reasons.

\begin{table*}
\footnotesize
\centering
\caption{Tentative structure for a taxonomy of XAI techniques.}
\begin{tabular}{lp{3.5cm}p{3.5cm}p{3.5cm}}
\toprule
& \multicolumn{2}{c}{Epistemic End} & \multicolumn{1}{c}{Means}\\
\cmidrule(lr){2-3} \cmidrule(l){4-4} 
& \emph{What?} & \emph{To whom?} & \emph{How?}\\
Example & (topic) & (stakeholder) & (instrument)\\
\midrule
\dots & \dots & \dots & \dots\\[2mm]
LIME & prediction(s) in specific part of the data & operators; executors & local approximation to complex model\\
&&&\\
CFE & input-output relation & decision-subjects & counterfactual statements\\
&&&\\
\citeauthor{Hendricks.2016} (\citeyear{Hendricks.2016}) & input-output relation & operators; executors & textual explanations\\
&&&\\
\citeauthor{Kim.2018} (\citeyear{Kim.2018}) & system's internal state & operators; executors & verbalized attention heatmapping\\[2mm]
\dots & \dots & \dots & \dots\\
\bottomrule
\end{tabular}
\label{tab:taxonomy}
\end{table*}


First, it reveals that although the XAI literature diverges both conceptually and methodologically, there is a common structure that is shared by the different approaches pursued in the literature. This structure consists in means-end relations that arise from the specification of topics, \textcolor{blue}{stakeholders}, and instruments.

Second, extending the analysis from above to other XAI techniques allows for a taxonomy of the existing literature in the field. Accordingly, it is possible to investigate the means-end relations governing other XAI techniques, to identify the relevant topic, \textcolor{blue}{stakeholder}, and instrument and to extend Table~\ref{tab:taxonomy} as indicated by the dots in the first and last row. Thus, the table is not only a description of four XAI techniques, it is also a starting point for a more comprehensive taxonomy of existing methods.

\subsection{The Account's Normative Component}
\label{subsec:normative}

Using the descriptive component of the means-end account, it is possible to unravel the different epistemic ends pursued by different authors and the different means they propose to achieve them. Apparently, authors come up with different instruments when they specify different topics or \textcolor{blue}{stakeholders}. Thus differences in the `what' and \textcolor{blue}{`to whom'} of an explanation seem to trigger differences in the `how' of achieving it. This observation might seem entirely unsurprising at first. After all, means-end considerations are common to a variety of contexts different from XAI in which instruments or techniques have to be chosen based on some given end: if I pursue the end of driving a nail into the wall, I better use a hammer instead of a violin bow. Yet if my end consists in practicing my favorite violin sonata, the latter will be much more appropriate.\footnote{\citet{GruneYanoff.2021} puts forward a similar argument for the case of scientific methodology.} The perspective of means-end epistemology allows to move beyond such intuitively plausible observations. In particular, it helps to make means-end relationships in XAI more precise, thereby ultimately leading to the normative component of the present account.

First, recall from \ref{eq:mee} that means-end epistemology is inherently normative and governed by the principle of instrumental rationality: given an epistemic end, certain means ought to be adopted if and only if they further the given end. Second, we have seen in \ref{eq:xai_mee} that problems of XAI can be framed as problems of means-end epistemology. In particular, specifying a topic and \textcolor{blue}{stakeholder} determines the epistemic end that is pursued, while specifying the corresponding XAI instruments determines the means to attain the given end. Taking these aspects together shows that XAI is inherently normative as well:

\ex. Given a topic $t$ and \textcolor{blue}{stakeholder $s$}, certain instruments of XAI ought to be adopted if and only if they further the epistemic end determined by $t$ and \textcolor{blue}{$s$}. \label{eq:nc}

Put differently, the suitability of XAI techniques depends on what exactly should be explained and \textcolor{blue}{to whom} the explanation should be given. However, this dependence is a normative one. Indeed, differences in the topic and \textcolor{blue}{stakeholder} determining an epistemic end trigger differences in the instruments to achieve it, but they do so with normative force. 

\textcolor{blue}{For instance, it is not by mere coincidence that the last column of Table~\ref{tab:taxonomy} displays four different instruments. Rather, the topics and stakeholders defined by the different authors determine different epistemic ends. The means-end account's normative component tells us that different ends call for different means. So clearly, given different topics or stakeholders, the specific instrument proposed by \citeauthor{Hendricks.2016} (\citeyear{Hendricks.2016}) differs from the one proposed by \citeauthor{Kim.2018} (\citeyear{Kim.2018}) which in turn differs from CFE. Furthermore, Table~\ref{tab:taxonomy} highlights the importance of analyzing epistemic ends at a high level of granularity. For instance, both the technique proposed by \citet{Kim.2018} and the one proposed by \citet{Hendricks.2016} are meant to provide explanations to either operators or executors of an ML system, yet the former authors aim at explanations of a system's internal states whereas the latter aim at explanations of its input-output relation. Thus, while addressing the same stakeholders, the techniques are meant to produce explanations of different topics, that is, of different aspects of an ML system. In sum, this leads to different epistemic ends being pursued and different means ought to be adopted to achieve them.}\footnote{\textcolor{blue}{The same reasoning could be applied when comparing the technique proposed by \citet{Hendricks.2016} to CFE. In that case, the topic is the same, but explanations should be produced for different stakeholders which, in sum, also leads to different epistemic ends (see Table~\ref{tab:taxonomy}).}}

Consequently, the examples hint at what the normative component allows to achieve on a more general level: it allows to spell out in great detail what an explanation for an ML method should `look like' in a specific setting. Thus, analyzing particular ends, that is, particular topics and \textcolor{blue}{stakeholders}, can reveal which instruments ought to be adopted to achieve them and which instruments can be ruled out as inappropriate. In order to allow for such insights, it is important to recognize the logical form in which the principle of instrumental rationality \ref{eq:mee} is stated in means-end epistemology and to preserve this form also in \ref{eq:nc}. Topic and \textcolor{blue}{stakeholder} constitute the epistemic end which is in turn tied by a biconditional to the adoption of certain means, that is, certain intruments of XAI: if certain instruments ought to be adopted, then they further the epistemic end determined by the given topic and \textcolor{blue}{stakeholder}; if they further the epistemic end determined by the given topic and \textcolor{blue}{stakeholder}, then certain instruments ought to be adopted. Considering only one of the latter conditional statements could give rise to two types of problematic cases: on the one hand, cases in which certain instruments ought \emph{not} to be adopted, although they further the given epistemic end.\footnote{There might be cases in which certain instruments \emph{cannot} be adopted, although they further the given epistemic end, for instance due to privacy issues or proprietary software. Yet this does not affect the normative result that, in principle, they nevertheless \emph{ought} to be adopted.} On the other hand, cases in which certain instruments do not further the given epistemic end, yet nevertheless ought to be adopted.

However, if such problematic cases are prevented, the normative component can be used to show how the set of appropriate XAI instruments is constrained by the particular ends for which an explanation is sought. As outlined above, in means-end epistemology and in the means-end account of XAI alike, the set of appropriate instruments gets narrower as the epistemic end gets more precise. 
This implies that one should aim for maximally specific topics and \textcolor{blue}{stakeholders} to identify a maximally specific set of appropriate instruments that one ought to adopt.\footnote{Note that no matter how specific topic and \textcolor{blue}{stakeholder} are defined, there will be several appropriate instruments and hence a set of different particular means to the same end in most cases (\citeauthor{Mothilal.2021} \citeyear{Mothilal.2021}).} Furthermore, the central insight of means-end epistemology that ``all we need to provide a means-end analysis [\dots] is a sufficiently clear description of the goals in question'' (\citeauthor{Schulte.1999} \citeyear{Schulte.1999}: 26) hints at the precondition for assessing the suitability of XAI techniques in a given context: we can only ever determine whether an XAI technique is appropriate, if \emph{ex ante}, the epistemic end being pursued is specified in sufficient detail. This implication of the present investigation is so far largely neglected in the discourse on XAI. In particular, high-level regulation such as the EU General Data Protection Regulation or the European Union's draft Artificial Intelligence Act does not spell out the precise epistemic end corresponding to the transparency requirements imposed on ML methods. 

Thus overall, the means-end account's normative component establishes a normative framework for the field of XAI. The framework is based on the insight that a close analysis of \emph{what} specific part of an ML method should be explained and \textcolor{blue}{\emph{to whom}} it should be explained allows to determine the set of instruments that is appropriate to bring about that specific explanation. This unifies and extends existing proposals that point to a similar direction. 

First, the means-end account's normative component unifies rather descriptive frameworks of XAI like the one proposed by \citeauthor{Sokol.2020} (\citeyear{Sokol.2020}). They carve out ``a set of descriptors that can be used to characterise and systematically assess explainable systems'' (\citeauthor{Sokol.2020} \citeyear{Sokol.2020}: 56). However, although leading to a useful and very detailed taxonomy of XAI techniques, the framework does not provide an overarching account of how the assessment of XAI techniques connects them to specific situations. This can be achieved using the means-end account's normative component: a specific situation will be characterized by a specific epistemic end being pursued which in turn normatively prescribes a set of appropriate instruments. For instance, consider a loan applicant---without expertise in ML---who asks for an explanation of why her loan application was accepted or rejected by the bank's ML-based decision system. The applicant's explanatory  topic---the decision, that is, the output of the ML-based system---and \textcolor{blue}{her own role as a specific stakeholder with specific explanatory requirements that need to be fulfilled} constitute an epistemic end that clearly rules out a highly specific explanation of the system's internal processes as the appropriate means. Instruments that provide some high-level explanation of why the system's input led to a specific output might be more appropriate in this case. Once such a set of appropriate instruments has been established, a purely descriptive account can help to assess whether a given XAI technique belongs to that set or not. Yet beforehand, it is epistemic normativity that links the characteristics of a situation to a set of instruments that are appropriate due to their specific characteristics and that hence ought to be adopted.

Second, the means-end account's normative component goes beyond other proposals of normative frameworks that solely focus on the explanatory requirements of different stakeholders. For instance, on \citeauthor{Zednik.2019}'s (\citeyear{Zednik.2019}) account, these requirements can be characterized by spelling out the epistemically relevant elements of an ML system for a given stakeholder. Since these elements differ across stakeholders, different explanatory requirements have to be satisfied by tailor-made explanations. Thus, an XAI technique should be used if it produces explanations that fulfill a stakeholder's explanatory requirements in virtue of addressing the specific epistemically relevant elements. However, fulfilling the requirements of different stakeholders can hardly be all there is to the suitability of XAI techniques, since one can easily imagine a situation in which the \emph{same} stakeholder asks for \emph{different} explanations. \textcolor{blue}{For instance, consider once more a loan applicant, but this time with a genuine interest in ML. In that case, she might on the one hand still ask for an explanation of why her loan application was accepted or rejected, that is, why certain inputs led to a certain output. On the other hand, however, she might also ask for an explanation of the ML system's internal functioning.} As already shown in Table~\ref{tab:taxonomy} above, the means-end account of XAI accomodates both situations as in this example and accounts that derive different explanatory requirements from differences across stakeholders in a straightforward way: on this account, the explanatory requirements of different stakeholders are but one source of variation in the epistemic end of XAI, another one consisting in the specification of different topics. \textcolor{blue}{The means-end account's descriptive component allows to distinguish them accurately and to analyze their particular specifications as well as the epistemic end that they determine. Additionally, given a certain epistemic end, the account's normative component allows to identify the set of instruments that the end calls for.}

\section{\textcolor{blue}{Extending the Account}}
\label{sec:extension}

\textcolor{blue}{We have made quite some progress up to this point: we have seen that the high-level aim of XAI can be reformulated in a more granular way (see (1')), that problems of XAI can be framed as problems of means-end epistemology (see \ref{eq:xai_mee}), and that, consequently, XAI is inherently normative (see \ref{eq:nc}). However, the preceding discussion exclusively focused on epistemic aspects, considering the epistemic end of producing explanations to be the main ingredient of the means-end account. Albeit leading to an account that accurately reflects the structure of XAI problems and is useful for their analysis, this perspective needs to be broadened for at least two reasons.}

\textcolor{blue}{First, there can be epistemic ends other than that of producing explanations. Indeed, it has been pointed out that sometimes, XAI techniques should produce explanations to achieve further epistemic ends such as understanding or interpretability \citep{Erasmus.2020}. Seen this way, achieving the epistemic end of producing explanations can also turn into a means to achieve another epistemic end.}

\textcolor{blue}{Second, XAI techniques are commonly employed to achieve non-epistemic ends. For instance, returning to the examples from above, LIME and the technique proposed by \citeauthor{Hendricks.2016} (\citeyear{Hendricks.2016}) are meant to foster trust in an ML system; \citeauthor{Kim.2018} (\citeyear{Kim.2018}) aim at enabling human agents to extrapolate the behavior of the vehicle controlled by an ML system; CFE are meant to inform, provide grounds to contest adverse decisions, and reveal how to reach a desired result in the future (\citeauthor{Wachter.2018} \citeyear{Wachter.2018}: 843).}

\textcolor{blue}{Thus, beyond the epistemic ends discussed above, there are further ends relevant to XAI that might stem from epistemic or non-epistemic considerations. So in addition to answering what should be explained, to whom it should be explained, and how it should be explained, researchers seem to answer the question as to \emph{why} something should be explained. I call the answer to this question the \emph{goal} of an explanation. Incorporating this into the means-end account, one can extend (1') by stating that XAI seeks to}

\ex.[(1*)] \textcolor{blue}{Provide instruments that produce explanations of topic $t$ for stakeholder $s$ to achieve goal $g$.} \label{eq:ext2}

\textcolor{blue}{This extension leads to a setup in which there are two types of means-end relations: the strictly epistemic means-end relations discussed above and means-end relations that can, but need not be strictly epistemic, depending on the specification of the goal. Importantly, the former relations are instrumental to the latter ones, since achieving the epistemic end of producing explanations of topic $t$ for stakeholder $s$ can also be a means to achieving goal $g$, provided that the particular epistemic end really is suitable to achieve the particular goal. This setup could easily be extended even further to what one might call a cascade of means-end relations in which achieving the end of one relation is also a means to achieve the end of the next relation and so on. To determine a suitable XAI technique, one would then have to work backwards through the cascade, ultimately arriving at one of the strictly epistemic means-end relations discussed above.\footnote{\textcolor{blue}{Apart from being intuitively plausible, this strategy of working backwards is suggested by the literature on means-end reasoning in both philosophy and artificial intelligence (\citeauthor{Bratman.1981} \citeyear{Bratman.1981}, \citeauthor{Pollock.1998} \citeyear{Pollock.1998}).}}}

\textcolor{blue}{Extending the means-end account in this way is important, since it reflects both the methodological and the conceptual literature on XAI. As for the methodological literature, consider once more the example of CFE. As discussed above, CFE are meant to reveal ``what could be changed to receive a desired result in the future'' (\citeauthor{Wachter.2018} \citeyear{Wachter.2018}: 843), a problem that is commonly referred to as \emph{algorithmic recourse} \citep{Venka.2020}. However, it has been claimed that CFE fall short of achieving this goal, for they ignore ``the \emph{causal relationships} governing the world in which actions will be performed'' (\citeauthor{Karimi.2021} \citeyear{Karimi.2021}: 353). This means that although CFE recommend a set of alternative actions (e.g., `increase annual income by amount $X$') these need not lead to the desired result (e.g., `you were granted the loan'), since the mechanism by which the ML model operates might not reflect the true underlying mechanism properly.\footnote{An intuitive example for this situation is provided in \citeauthor{Karimi.2021} (\citeyear{Karimi.2021}: 353).} The debate surrounding CFE reveals that, although implicitly, debates about the suitability of XAI methods indeed center around means-end considerations and their inherent normativity: according to the literature, counterfactual statements are an appropriate means to achieve the epistemic end of producing explanations of an ML system's input-output relation for a decision-subject (see Table~\ref{tab:test}). Furthermore, this epistemic end seems to be an appropriate means to achieve the goal of identifying and contesting adverse decisions. However, it seems to be an inappropriate means to achieve the goal of providing information on how to turn them into a desired result in the future. Consequently, CFE ought not be used when the latter is the goal.}\footnote{Similarly, it has been questioned whether producing particular explanations is an appropriate means to achieve the goal of establishing human trust in an ML system \citep{Kastner.2021}.}

\textcolor{blue}{As for the conceptual literature, recall from above that \citeauthor{Krishnan.2020} (\citeyear{Krishnan.2020}: 488) argues that what is called explainability or interpretability ``serves as a means to an end, rather than being an end in itself'' (\citeauthor{Krishnan.2020} \citeyear{Krishnan.2020}: 488).\footnote{Similar arguments have been put forward, e.g., by \citet{Bordt.2022}.} Thus, beyond the basic epistemic end of producing explanations and beyond further epistemic ends such as interpretability, there are more `fundamental goals' (\citeauthor{Krishnan.2020} \citeyear{Krishnan.2020}: 495) that she takes to be the \emph{actual} ends that should be achieved. Accordingly, problems of XAI should be framed in a way that uncovers these fundamental goals to ``facilitate a more pluralistic approach to problem-solving'' (\citeauthor{Krishnan.2020} \citeyear{Krishnan.2020}: 495). So on her view, identifying XAI techniques that are appropriate in a particular situation requires a fine-grained characterization of that situation in the first place. The extension introduced in (1*), distinguishing epistemic ends from further goals that might be specified, offers a straightforward way to accommodate this line of argumentation within the means-end account of XAI. First, we have seen that epistemic ends come in varying degrees of granularity. Second, the account's normative component reveals that more or less specific ends entail more or less specific sets of appropriate instruments. Third, different (sets of) means ought to be adopted to achieve different epistemic ends. Fourth, different epistemic ends are required to achieve different goals. Consequently, the means-end account of XAI allows for a fine-grained investigation of what Krishnan refers to as `fundamental goals'. It also reveals why a pluralistic approach to problem-solving is indeed necessary in XAI: since one ought to adopt the means to one's ends.}

\textcolor{blue}{Having said all this, however, one might object that the perspective of normative epistemology adopted in this article is no longer essential to the means-end account. After all, what matters when faced with a cascade of means-end relations seems to be the strategy of backward-planning mentioned above, since this suffices to ensure means-end coherence across the different relations. But we do not have to consider an entire cascade of means-end relations, we do not even have to go beyond (1*) to notice the following: XAI is an epistemic endeavor at heart. No matter what further goals are specified, we have seen that a strictly epistemic means-end relation is the nucleus of everything that follows. This is why the perspective of normative epistemology is crucial.}

\section{Conclusion}
\label{sec:conclusion}

This paper set out by observing considerable disagreement in the XAI literature, both on a conceptual and on a methodological level. Does this mean that the field of XAI is scattered into various disconnected subprojects? Quite to the contrary. Indeed, I argue that there is a common structure that is shared by the variety of approaches pursued in the literature. To do so, I put forward a means-end account of XAI. The account relies on the insight from means-end epistemology that one ought to adopt appropriate means to one's epistemic ends. It also relies on the observation that XAI can be framed as a problem of means-end epistemology. Taking both aspects together, I show that normative means-end relations are and should in fact be central to XAI. I further show that these means-end relations are determined by a topic, a \textcolor{blue}{stakeholder}, a goal, and an instrument. By specifying these, one provides answers to four central questions: What should be explained? \textcolor{blue}{To whom} should it be explained? Why should it be explained? And how should it be explained? 

The means-end account of XAI has several important consequences. First, it explains why disagreement arises in the field: the divergence in instruments of XAI follows from the disagreement on epistemic ends. Second, it structures the field: there is a common methodology of developing appropriate means for given ends. Third, this structure has a descriptive component: different authors specify different ends and come up with different means to achieve them. This gives rise to a taxonomy that classifies existing contributions to the field along the specific means-end relations that are considered. Fourth, this structure also has a normative component: the ends of an explanation normatively constrain the set of admissible means to achieve it. The means-end account thus reveals how the suitability of particular instruments of XAI is prescribed by the ends for which an explanation is sought.

Future research might investigate the different components of the means-end account even further. On the one hand, I plan to establish a more comprehensive taxonomy of existing XAI techniques using the account's descriptive component and to evaluate it using the normative component. On the other hand, there is potential to apply the means-end account to regulatory issues. From this perspective, the account paves the way for designing effective guidelines of XAI that ensure that the right techniques are employed in the right situations. 

\clearpage



\end{document}